\documentclass[fleqn,10pt,dvipsnames]{wlscirep}
\usepackage[utf8]{inputenc}
\usepackage[T1]{fontenc}

\usepackage{lineno}
\usepackage[colorinlistoftodos]{todonotes}
\usepackage{soul}

\title{MIDAS: Deep learning human action intention prediction from natural eye movement patterns}

\author[1,2,3,4,+]{Paul Festor}
\author[1,2,3,+]{Ali Shafti}
\author[1,3]{Alex Harston}
\author[1,2]{Mickey Li}
\author[1,2,3]{Pavel Orlov}
\author[1,2,3,4,5,*]{A. Aldo Faisal}
\affil[1]{Brain and Behaviour Lab: Dept. of Bioengineering, Imperial College London, SW7 2AZ, London, UK}
\affil[2]{Dept. of Computing, Imperial College London, SW7 2AZ, London, UK}
\affil[3]{Behaviour Analytics Lab, Data Science Institute, SW7 2AZ, London, UK}
\affil[4]{UKRI CDT in AI for Healthcare, Imperial College London, SW7 2AZ, London, UK}
\affil[5]{MRC London Institute of Medical Sciences, W12 0NN, London, UK}
\affil[*]{corresponding author: A. Aldo Faisal (aldo.faisal@imperial.ac.uk)}
\affil[+]{these authors contributed equally to this work}

\begin{abstract}
Eye movements have long been studied as a window into the attentional mechanisms of the human brain and made accessible as novelty style human-machine interfaces. However, not everything that we gaze upon, is something we want to interact with; this is known as the Midas Touch problem for gaze interfaces. To overcome the Midas Touch problem, present interfaces tend not to rely on natural gaze cues, but rather use dwell time or gaze gestures. Here we present an entirely data-driven approach to decode human intention for object manipulation tasks based solely on natural gaze cues. We run data collection experiments where 16 participants are given manipulation and inspection tasks to be performed on various objects on a table in front of them. The subjects' eye movements are recorded using wearable eye-trackers allowing the participants to freely move their head and gaze upon the scene. We use our Semantic Fovea, a convolutional neural network model to obtain the objects in the scene and their relation to gaze traces at every frame. We then evaluate the data and examine several ways to model the classification task for intention prediction. Our evaluation shows that intention prediction is not a naive result of the data, but rather relies on non-linear temporal processing of gaze cues. We model the task as a time series classification problem and design a bidirectional Long-Short-Term-Memory (LSTM) network architecture to decode intentions. Our results show that we can decode human intention of motion purely from natural gaze cues and object relative position, with $91.9\%$ accuracy. Our work demonstrates the feasibility of natural gaze as a Zero-UI interface for human-machine interaction, i.e., users will only need to act naturally, and do not need to interact with the interface itself or deviate from their natural eye movement patterns. 
\end{abstract}
\begin{document}

\flushbottom
\maketitle

\thispagestyle{empty}

\section*{Introduction}
The way we interface humans with machines can define the success of their interaction, and is in many cases the bottleneck for overall performance \cite{Laurel1990}. Human eye movements have long been studied as a window to human cognition, with the resulting gaze-based interfaces seen as a promising medium for low-cognitive-load human-machine interaction \cite{abbott_ultra-low-cost_2012}. Interfaces of this kind aim to infer intention from the way in which a user looks at the world\cite{adjouadi2004remote,chin2008integrated}. A limiting factor to their development however is that they suffer from the `Midas Touch' problem -- the problem of distinguishing whether a user is simply viewing an object to intake visual information about it, or actively viewing because they intend to interact with it \cite{velichkovsky1997towards}. The Midas Touch problem remains unsolved in gaze-based interfaces, due to the challenges within accurate eye-tracking, in variable environmental conditions, resulting in imperfect data \cite{Niehorster2020} as well as the complexity of interpreting gaze dynamics in an ever-shifting visual environment and mapping these to a range of possible intentions\cite{Velichkovsky2014-da}. 

Researchers have studied gaze as a proxy for human intentions for over a century \cite{Buswell1935-hf, yarbus1967eye, grasso_eye-head_1998, hollands_look_2002}, predominantly with respect to the major constituent parts of gaze, i.e. saccades and fixations \cite{Becker1969-yj, Fuchs1976-vf}. Measures of intention in this regard have generally focused either on differences in the distribution of gaze fixations (defined as periods of comparatively little movement, for time periods of around 350ms \cite{Rayner1992-tv}), or on differences in saccadic velocity profiles \cite{jacob_what_1990, salvucci2000identifying, duchowski_breadth-first_2002}. Since the days of Yarbus \cite{yarbus1967eye} the field has shown spatial differences in gaze scanpaths in different static task contexts. Such paradigms of showing a stimulus and observing a response lie at the heart of vision science \cite{Lorenz2016-es}; but here we aim to invert this traditional approach using machine learning, to investigate how we can decode the intentions of a person by observing \emph{how} they look at a scene, based on purely natural gaze traces. Such an approach would unlock not only the use of natural gaze for human interfacing by enabling us to overcome the Midas Touch problem but would also unlock our scientific understanding of what controls visual saliency and task goals.

Many prior works have gone beyond static fixation distributions and investigated gaze behaviour in natural settings \cite{Rothkopf2004-wt, land_making_tea_1999, Hayhoe2003-be, keshava2021just, Johansson2001-mj}, to try and capture the implicit information held within natural gaze. For `reverse engineering' gaze behaviour to understand intention, there have been attempts at building Inverse Yarbus models\cite{Haji-Abolhassani2014-gf}, though most eye movement research is based on participants looking at a computer monitor with a static image \cite{Greene2012-ka, Borji2014-py, Iqbal2004-vz, Henderson2013-oz}. Approaches like these may be suitable for modelling simplistic static gaze behaviour where multiple saccades and fixations are made over a constant image, but are not applicable for more complex free-viewing behaviours.  The efficacy and applicability of such Inverse Yarbus models outside of very limited static scenarios has been contested \cite{Castelhano2009-mq, Greene2012-ka, Borji2014-py}. With the exception of Bulling et. al's work \cite{Bulling2013-qv}, which focuses on a simplified binary classification problem, no Inverse Yarbus models detailed in the literature have been developed to work in real-world settings -- that is, no models have been trained using data gathered from real-world natural experiments, where participants freely move their heads and interact with real physical objects in 3D \cite{Boisvert2016-my}. One of the reasons for this is the increased difficulty in overcoming the Midas Touch problem in natural settings -- we do not necessarily want to interact with every object we look at, and in object-dense real-world environments this problem is magnified significantly.  

Past efforts to overcome the Midas Touch problem have involved techniques such as analysis of dwell-time \cite{Hansen2003-su}, detection of focal fixation \cite{Velichkovsky2014-da}, or co-actuation with other modalities, such as voice \cite{Parisay2020-xw}, winking \cite{dziemian2016gaze}, or keyboard input \cite{drone_GCI_2014} but such attempts are only successful in limited and simplified settings. These methods have the advantage of correctly interpreting the user's intentions, but ignore all natural gaze cues and force end-users to direct their gaze in an artificial manner, which can prove both difficult for users, and unreliable in all but the most narrow of scenarios. 

Co-actuation of gaze with a voluntary movement is also not suitable for those users who would be the most direct beneficiaries of natural gaze interfaces, i.e. users with severe motor impairments. For these users, many of the typical human-machine interfaces which require actuation by limbs or digits, are not usable. The human oculomotor system typically remains intact even in severe cases of tetraplegia and other motor impairments\cite{abbott_ultra-low-cost_2012}, and eye movements are preserved for a longer period than skeletal movement with respect to neurodegenerative diseases \cite{tostado20163d}, making them a viable interface for disabled users. Ideally, a gaze interface would rely solely on natural gaze cues, such that the user would direct their gaze as they naturally do, with the system having the capability to detect action intention from subtle differences in the natural gaze signal, thereby minimising cognitive load. In addition, natural gaze contains temporal characteristics that models can leverage, such as a sequential structure \cite{land_making_tea_1999, Johansson2001-mj} and just-in-time order \cite{Hayhoe2003-be, keshava2021just} when actively viewing freely in natural tasks.  Hidden Markov Models have already been applied to scanpath modelling \cite{Coutrot2018-jn} for predicting future actions based on gaze information, however these models suffer considerably when data is noisy and incomplete, as is often the case in real-world recordings, and are unreliable in complex situations with more flexible behaviours. 

\begin{figure}[htbp]
\centering
\includegraphics[width=0.9\linewidth]{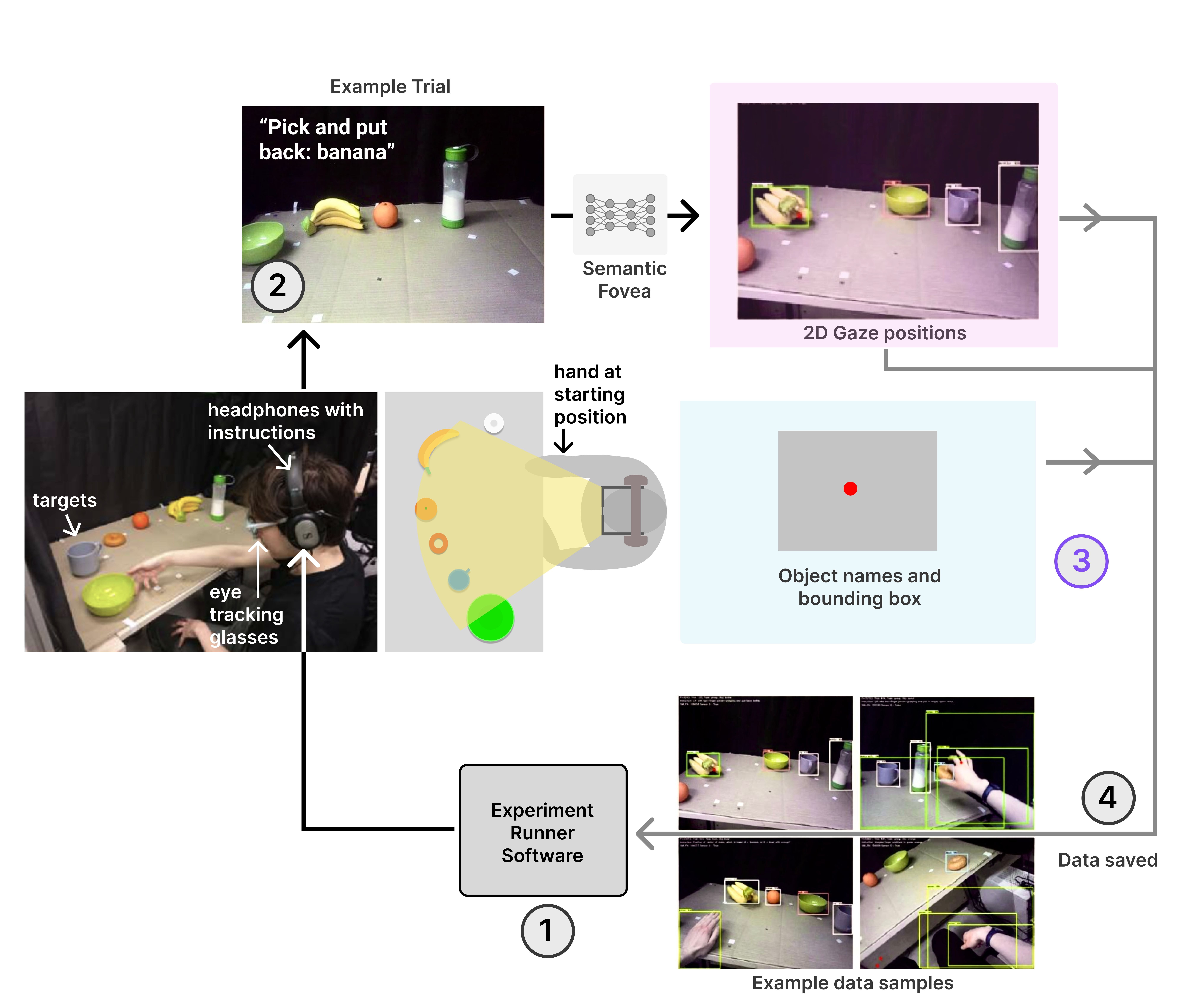}
\caption{Data collection experiments. (1) Our experiment runner software manages the experiment, randomising the tasks and keeping track of progress. The software produces voice commands, asking the user to perform tasks requiring physical/imaginary manipulation of objects or inspection of attributes of the objects on the table. These commands are communicated to human participants through headphones (left). The participants are seated at the table with the objects placed on it, and are wearing eye-tracking glasses that record their eye-movements, and an ego-centric video stream. The subjects will then proceed to fulfil the task given to them by the experiment runner. An example of this can be seen in an ego-centric view in (2). The ego-centric video frames are passed through our Semantic Fovea\cite{auepanwiriyakul2018semantic_fovea} which detects objects in the scene and their position with respect to gaze. The generated data, i.e. 2D gaze positions, object names and bounding box positions are then all saved by the experiment runner software (4). All frames shown in the example trial, 2D gaze positions and example data samples subfigures are egocentric views recorded from our dataset.}
\label{fig:data_collection}
\end{figure}

Here we detail a new approach by capturing multimodal data sources from real-world human behaviour. By combining spatiotemporal eye movement dynamics with contextual information extracted on-the-fly through the implementation of a `semantic fovea' (an object recognition system that can recognise and categorise objects in the field of view in real-time \cite{auepanwiriyakul2018semantic_fovea}), we obtain rich natural gaze behaviour, allowing our models to learn and distinguish subtle differences in the user's gaze dynamics. We designed an experimental process to capture task-specific natural gaze behaviour data in a dining table scenario (see Fig.~\ref{fig:data_collection}). Subjects sat in front of a dining table where different target objects were placed. Each subject was asked, through computer-generated voice commands, to perform actions of three types: manipulate one of the objects, imagine manipulating one of them, or inspect objects to answer a question. Experiments were conducted with 16 subjects, for about 3 hours of interactions each, resulting in $13,679$ total trials. 

This multimodal real-world experimental approach allows us to gather sufficient quantities of high resolution contextual gaze data, such that we can build deep recurrent neural network models that can capture and distinguish subtle differences in the temporal and spatial evolution of scanpaths over salient objects. Our models can thereby accurately identify task context, based solely on a participant's stream of eye movements in a given real-world task, providing a promising step forward for data-driven natural gaze control interfaces. 

\section*{Results}

\subsection*{Dataset description}
We ran data collection experiments with 16 subjects, for approximately 3 hours each. During each experiment trial, subjects interacted with 6 different target objects (a banana, a bottle, a bowl, a cup, a doughnut, and an orange) on a table in front of them,  see Fig.~\ref{fig:data_collection}. Interactions were prompted by instructions communicated through computer-generated voice commands. Each instruction was composed of what to do and which target object to do this on e.g., \emph{``Pick and place in the center: banana''}. See supplementary materials for a comprehensive list of all instructions. The instructions covered two classes of behaviour: \emph{inspection}, where users would visually inspect objects' attributes to answer a two-choice question, and \emph{manipulation} where users would manipulate an object, either in physical or imaginary interaction. The structure of each instruction sentence was designed to reveal the target of the task to the subjects at the very end of the command, to minimise task-object related gaze cues appearing earlier than the end of an instruction communication. 

A behavioural dataset was collected through these experiments. Participants wore a pair of eye-tracking glasses (SMI ETG 2W, SensoMotoric Instruments, Germany), which recorded both ego-centric video and gaze position in pixels within the video frame, at average frequencies of 30Hz and 120Hz, respectively. Video frames were post-processed, going through our Semantic Fovea \cite{auepanwiriyakul2018semantic_fovea}, a convolutional neural network trained for object detection within ego-centric videos, relative to gaze points, producing object names and bounding boxes, see Fig.~\ref{fig:data_collection}. Our experimental setup was equipped with optical sensors, to monitor which object slots were occupied, and for the subjects to answer the two choice question on object attributes presented to them in each inspection task by placing their hand on one of two sensors. See Methods for details. The sensor values are part of the dataset but were not used for our classification study. The intention decoding models we present here take as features the gaze and object positions within a video frame. These signals were not recorded at the same frequency; gaze points were recorded at 120Hz while object positions were extracted from video frames recorded at 30Hz. Moreover, the object detection algorithm can occasionally miss objects in a given frame. We used interpolation to make up for the gaps resulting from recording frequency mismatches and object detection errors. See Methods for details. 

\begin{figure}[htbp]
\centering
\includegraphics[width=\linewidth]{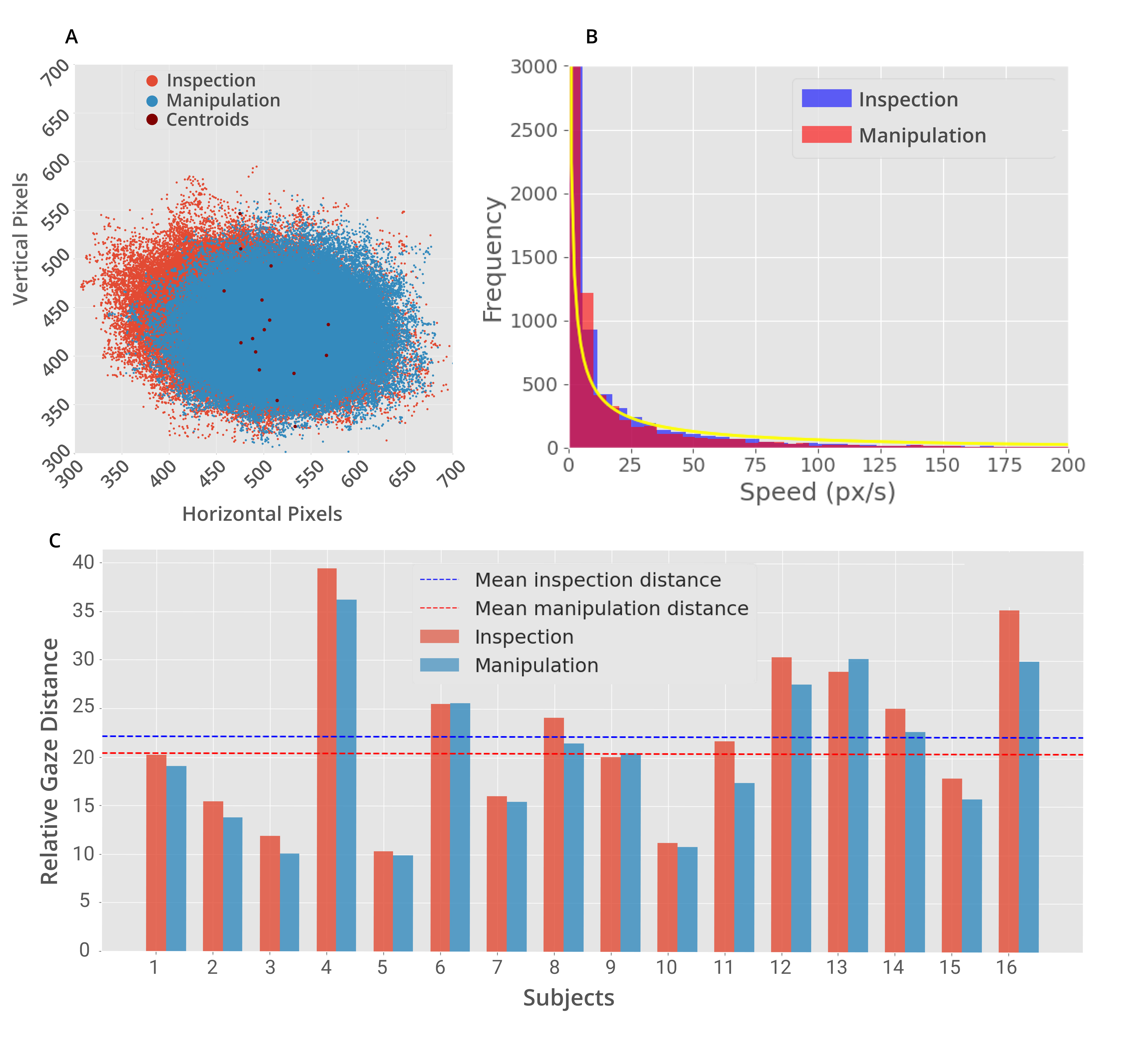}
\caption{Dataset overview. A) shows a slight difference in the mean 2-D distributions of gazepoints across all subjects in the two different tasks, with the manipulation task right-shifted across all subjects. We also overlay in purple the centroids of the subjects' individual gaze distributions on the grand average per-task distributions. B) shows a histogram of the speed distribution difference between the two different modalities, both fitting a chi squared distribution. C) Mean
gaze distance covered over all subjects’ respective trials, showing variance on a per-subject basis as considerably
higher than the difference between the two tasks}
\label{fig:dataset_oveview}
\end{figure}

The experiment was designed to be conducted in 1-hour batches of trials, with 10 minutes of break every hour. In total, 16 subjects took part in the experiments, all right-handed, completing two or three 1-hour long batches each, resulting in a total of $13,679$ trials, evenly split between manipulation (itself split into physical or imaginary, $53.4\%$ and $46.6\%$ respectively) and inspection tasks. See Fig.~\ref{fig:data_collection} for an overview of the experimental procedure. Gaze data recorded from all 16 subjects follows normal Gaussian distributions both in the horizontal and vertical axes of the egocentric image frame, with the emergence of a central bias \cite{Tatler2007-gr} of gaze behaviour in head-centric coordinates, see Fig.~\ref{fig:dataset_oveview}A. There is slight but noticeable difference when comparing the centrally-biased distributions of the two classes of tasks, with the manipulation task showing a consistent $\approx$100 pixel offset in the horizontal axis across all subjects, and a negligible difference in the vertical axis (Fig.~\ref{fig:dataset_oveview}A).

Gaze speeds across all subjects follow similar patterns, showing a chi-squared distribution (Fig.~\ref{fig:dataset_oveview}B).
When comparing mean gaze distance covered over all subjects' respective trials, we can see that the variance on a per-subject basis is considerably higher than the difference between the two tasks (Fig.~\ref{fig:dataset_oveview}C). We note however that there is a small but systematic difference in the mean gaze distance covered between inspection and manipulation trials, with the distance covered across inspection trials systematically higher than that of manipulation trials (Fig.~\ref{fig:dataset_oveview}C). We investigate these systematic differences in the analysis and validation of our classifier.

\subsection*{Gaze-based intention classification}
\begin{figure}[htbp]
\centering
\includegraphics[width=\linewidth]{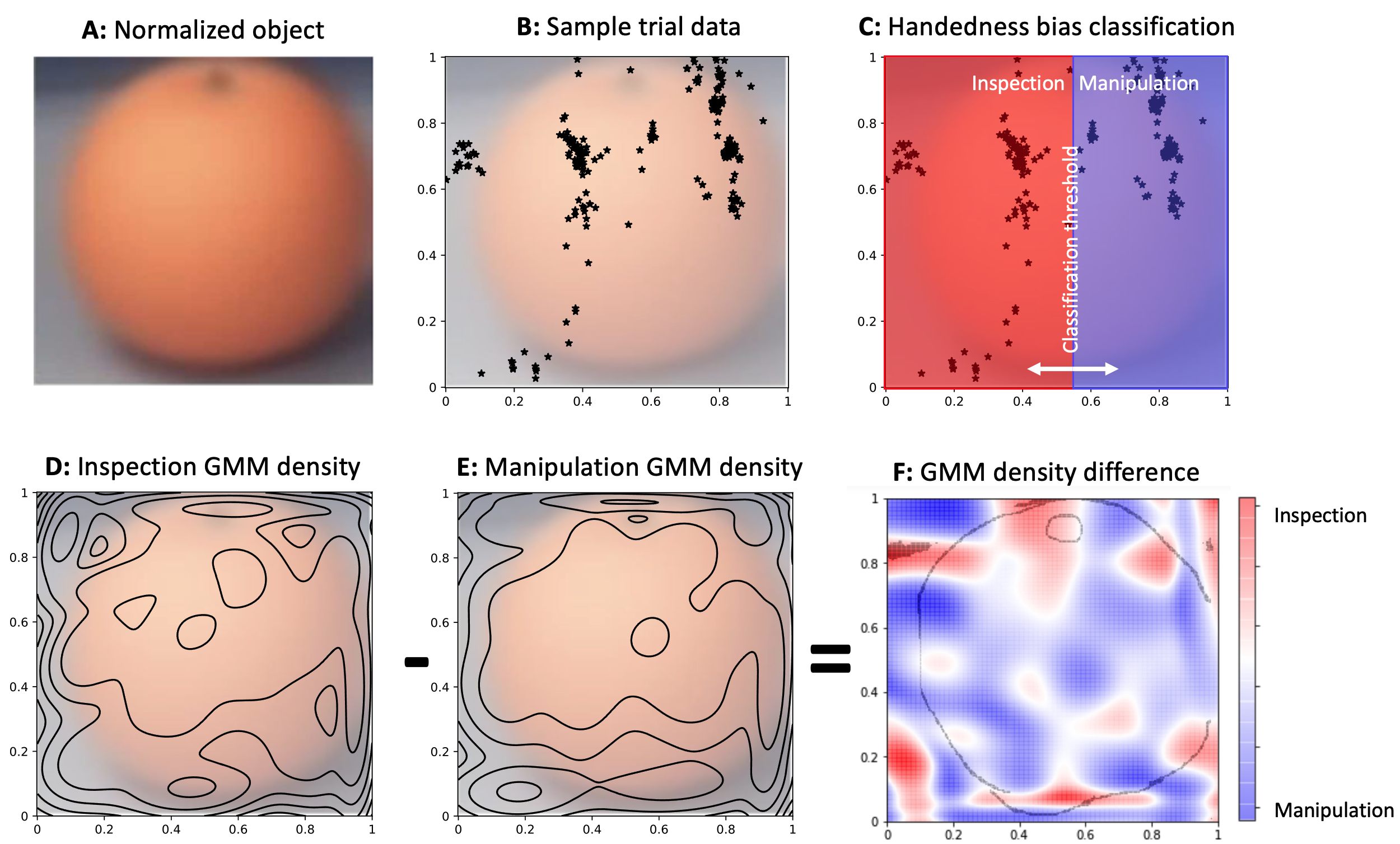}
\caption{Illustration of the handedness bias and GMM intent classifiers with the orange as an example target object.
\textbf{(A)} Normalised picture of the orange.
\textbf{(B)} Gaze points landing on the orange on a sample trial where it is the target object.
\textbf{(C)} Illustration of the handedness bias classifier, which looks at the relative amount of eye movements to the left or right of the vertical decision boundary between the red section (inspection) and the blue section (manipulation) - for a right handed user using the right hand for interaction.
\textbf{(D)} Contour plot of the learnt GMM density for the inspection trials.
\textbf{(E)} Contour plot of the learnt GMM density for the manipulation trials.
\textbf{(F)} Heatmap of the  difference between the learnt GMM densities for inspection and manipulation trials. 
}
\label{fig:Linear_and_GMM}
\end{figure}

As a first model for predicting gaze context from scanpaths, we investigate the horizontal shift of gaze points observed in the data between the two types of tasks (see Fig.~\ref{fig:dataset_oveview}A). We focus on static 2D gaze positions on the target object, attempting to classify intentions based on the average horizontal gaze position within the target object bounding box. 
Fig.~\ref{fig:Linear_and_GMM}A shows the average normalised image of an orange obtained by normalising all the orange object frames given by the Semantic Fovea\cite{auepanwiriyakul2018semantic_fovea}.

The naive hypothesis here is that as all our participants are right-handed, and would reach to manipulate objects from the right-hand side, the distribution of gaze positions within the object bounding box would show a higher concentration on the right-hand portion, where they would place their hand -- we refer to this as handedness bias. To check for this hypothesis, we implemented our handedness bias classifier which takes as input the average of the horizontal positions of the gaze points on the object, with gaze positions normalized to the object bounding box, and classifies the user’s intention as manipulation if this average is above a given threshold. The threshold could vary from $0$ (left border of the object) to $1$ (right border of the object). If the average normalized horizontal gaze position is above the threshold, then the trial is considered as manipulation, otherwise it is identified as inspection.

Fig.~\ref{fig:Linear_and_GMM}B shows data from a sample trial where the target object was the orange and Fig.~\ref{fig:Linear_and_GMM}C shows a sketch of the handedness bias model. This model was evaluated with thresholds ranging from $0$ to $1$ by steps of $0.01$ and with best accuracy reached on our dataset being $51\%$, i.e. almost chance level. Such low accuracy on a balanced binary classification task indicates that the average horizontal gaze position on the object does not carry distinguishable information with regards to the context of gaze, i.e. the intended task, and is therefore not a good feature to infer the user’s intention from.

For a more general evaluation of gaze position on target object as a feature for intention decoding, we set out to determine whether the distribution of gaze positions on the target object differs from one intention to the other. We used Gaussian Mixture Models (GMMs) \cite{reynolds_GMM_2009} to learn the static distribution of gaze positions on the target object for each intention class. A GMM is a flexible model for probability distributions, which fits a weighted set (or mixture) of Gaussian distributions to a target distribution. Here we are trying to model two target distributions, one per intention, so we fit two different GMMs to our data. To classify a trial, the set of gaze points on the target object is extracted, and the likelihood of this set under each of the models is computed. The intention with the highest likelihood is then associated to the observed gaze data. The number of Gaussians in each GMM was determined by optimizing the Bayesian Information Criterion (BIC) \cite{neath_BIC_2012}. 

Fig~\ref{fig:Linear_and_GMM}E and Fig~\ref{fig:Linear_and_GMM}F show the learnt distributions for inspection and manipulation trials respectively, on an orange. To help in reading the difference, Fig~\ref{fig:Linear_and_GMM}F shows the difference between the two learnt models. The accuracy of the GMM approach in decoding intention was $53\%$, barely above chance level. This result indicates that static gaze distributions on the target object do not carry enough information to discriminate between user intentions. Two factors limit the approaches we have shown here: ignoring the time dimension, and the dismissal of all gaze positions which did not land in the target object’s bounding box.

\begin{figure}[htbp]
\centering
\includegraphics[width=\linewidth]{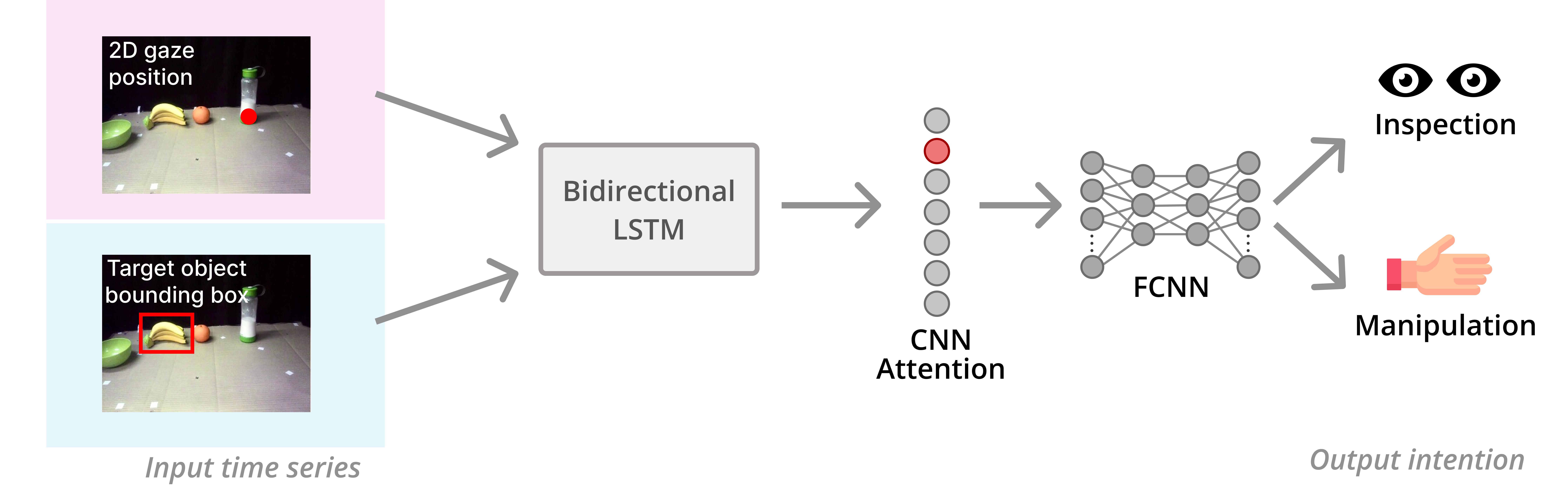}
\caption{Sketch of MIDAS's architecture. Inputs are 2D gaze and target position time series (or any other time series feature such as gaze speed or gaze-target distance), color coding refers to Fig.~\ref{fig:data_collection}. The input time series are fed to a bidirectional LSTM layer, an attention layer and finally a fully connected network which outputs the intention class.}
\label{fig:model_sketch}
\end{figure}

\begin{figure}[htbp]
\centering
\includegraphics[width=\linewidth]{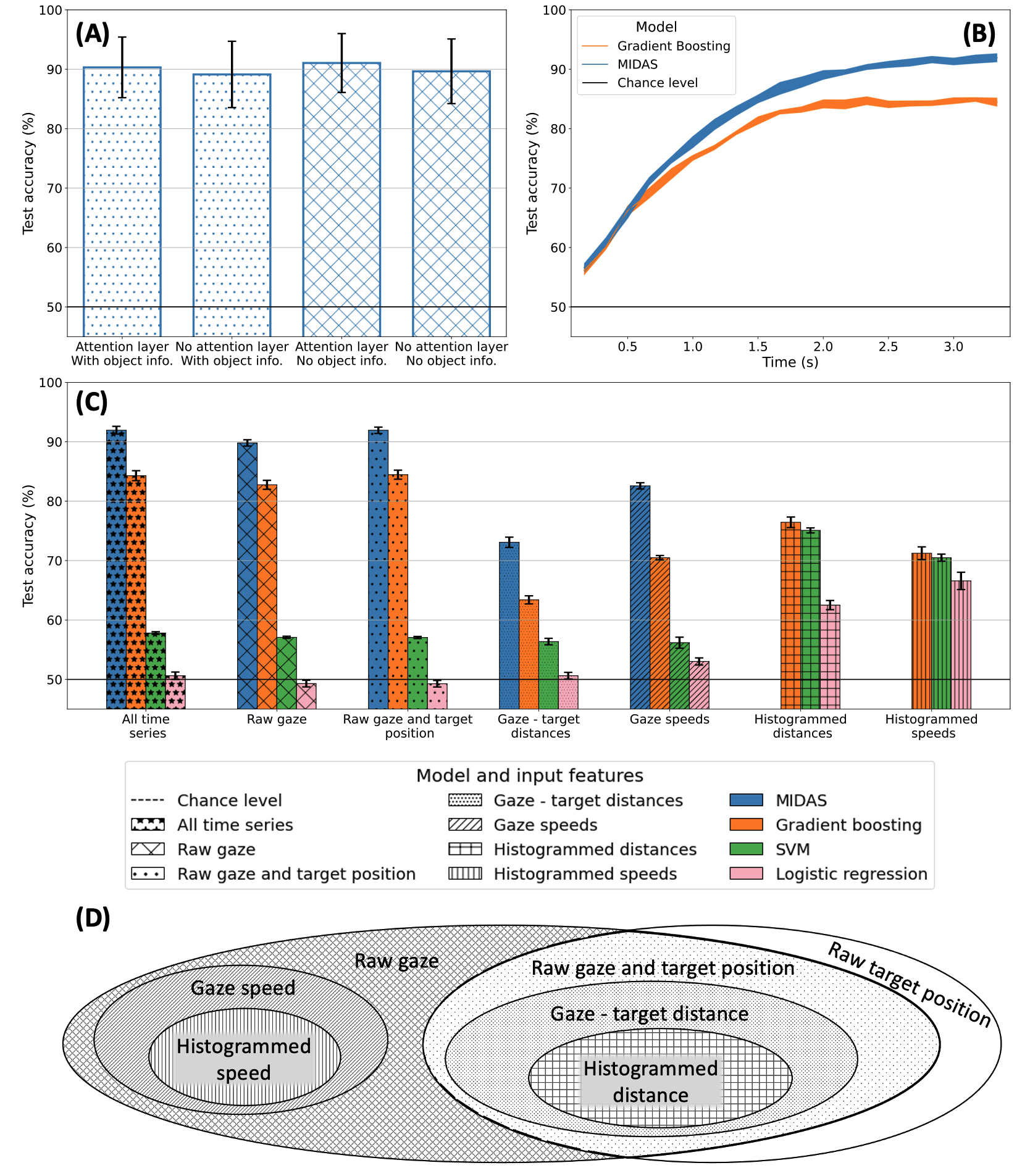}
\caption{The performance of our model, MIDAS, in ablation studies and in comparison with standard models.
(a) Mean and standard deviation of MIDAS's accuracy on leave-one-subject-out cross-validation, with ablations on inclusion of the attention layer in the deep network, and the target object position.
(b) Accuracy of MIDAS, gradient boosting and random forest over time from the moment when the subject has heard the full task description.
(c) The performance of MIDAS on 5-fold cross-validation compared to more standard machine learning classification models. In all plots, the dashed black line indicates chance level.}
\label{fig:model_results}
\end{figure}

The observations from the results above confirm findings described in the literature regarding the importance of gaze dynamics and the order of motion in gaze scanpaths over time \cite{land_tea_1999}. To incorporate this dimension into our classifiers, we implemented models which use time series data as input. We considered different sets of features: gaze position, gaze and target object positions, gaze-target distance and gaze speed. We also trained models to classify intention from gaze-target distances distribution and gaze speeds distribution. See Fig.~\ref{fig:model_results}D for a summary of the different features considered. The purpose is now to find a model which does well at classifying the different input sequences to recognize the user’s intent. 

Treating intention decoding as a sequence classification problem is well studied in Natural Language Processing (NLP), particularly in the task of sentiment analysis, where algorithms are designed to decode the main sentiment of a sentence \cite{rosenthal_semeval-2017_2017}. Here, we took inspiration from the work of Baziotis et. al.\cite{baziotis_datastories_2017} who have used an LSTM-based sentiment decoder running on tweets. Fig.~\ref{fig:model_sketch} presents the architecture of our gaze intention classifier, MIDAS. First, input time series (2D gaze position alone or with target object position, gaze speed or gaze-target distance) are fed to a bidirectional LSTM network \cite{graves2005framewise_bidirLSTM}, which captures the time dynamics of the input features. The output of the LSTM block is then fed to an attention layer \cite{bahdanau_neuralAttention_2014} which focuses the model on the most meaningful chunks of the time series. The output of the attention layer is fed into a fully connected neural network which outputs the user’s intention. To benchmark MIDAS’s performance, we also trained standard machine learning models to decode user intention from our behavioral gaze dataset, namely Gradient Boosting, Support Vector Machines (SVMs), and Logistic Regression.

Different feature sets are explored to classify intent. Given its architecture, MIDAS only allows time series data as input (2D gaze positions, gaze-target distance or gaze speed, see Fig.~\ref{fig:model_results}D). All other machine learning models used in this study only require fixed-size inputs, meaning that they accept both fixed-length time series and their histograms. The models were trained and evaluated under cross-validation schemes, in which the available data is split into several chunks, and the model is trained successively on all except one left out for validation. We used two cross validation schemes: 5-fold where the full dataset is randomly split into 5 chunks of data, and leave one subject out where each chunk is the data from one subject. The latter allows to evaluate how well the model would generalize when being fed data from new, unseen users. 

Fig.~\ref{fig:model_results}A shows the mean and standard deviations of MIDAS’s accuracy from leave one subject out cross-validation. The patterns on the bars in Fig.~\ref{fig:model_results}A correspond to those shown in the Venn diagram in Fig.~\ref{fig:model_results}D. The highest mean accuracy reached by MIDAS on leave one subject out cross-validation is $91.1\%$, for the case where an attention layer is used in the architecture, and no target object position information is provided. However, the MIDAS accuracy does not vary much without attention, or with inclusion of target object position data, as can be seen in Fig.~\ref{fig:model_results}A, reaching a minimum of $89.13\%$ mean accuracy in the case with no attention layer and no target object position data. This indicates that the result is robust to the inclusion/exclusion of an attention layer in the architecture, and also that knowing the target object a priori is not a requirement to maintain the model's performance level.

Fig.~\ref{fig:model_results}C shows means and standard deviations for MIDAS and standard machine learning approaches, on 5-fold cross validation, with different feature set inputs; here as well, the bar chart patterns are indicative of feature type and correspond to those shown in the Venn diagram in Fig.~\ref{fig:model_results}D. Given raw gaze traces and the target object's position as input MIDAS outperforms standard machine learning approaches on the intention decoding task, rising up to $91.9\%$ mean accuracy, compared to highest mean accuracy of other models being $84.5\%$ in the case of gradient boosting. The same observation can be made with gaze speed  and gaze-target distance time series as input, where MIDAS reaches respectively $82.6\%$ and $73.1\%$ accuracy, each at about $10\%$ above the next best performing model. Gradient boosting outperforms logistic regression and SVM on this task, no matter the feature set. It can also be noticed that the histogrammed feature sets carry some information about the user's intent as classifiers can reach an accuracy significantly above chance with them. Table \ref{tab:5_fold_accuracies} shows detailed results of the classification accuracy for all models on 5-fold cross validation.

\begin{table}[htbp]
    \centering
    \begin{tabular}{|l|c|c|c|c|c|}
    \hline
        Input Feature & MIDAS & Gradient boosting & SVM & Logistic regression & Handedness bias\\
        \hline
        Raw gaze & $\mathbf{89.8 \pm 0.5}$ & $83.1 \pm 0.6$ & $56.9 \pm 0.9$ & {\color{gray}  $49.7 \pm 0.2$} & N/A\\
        \hline
        Raw gaze and target position & {\color{Green} $\mathbf{91.9 \pm 0.5}$} & $84.5 \pm 0.6$ & $56.8 \pm 0.9$ &{\color{gray}  $49.7 \pm 0.2$} & {\color{gray} 
        $50.5 \pm 0.1$ }\\
        \hline
        Gaze - target distances & $\mathbf{73.1 \pm 0.9}$ & $63.7 \pm 0.7$ & $56.1 \pm 1.2$ &{\color{gray}  $50.5 \pm 1.0$ }& N/A\\
        \hline
        Gaze speeds & $\mathbf{82.6 \pm 0.5}$ & $70.8 \pm 1.0$ & $56.4 \pm 0.8$ & $52.9 \pm 0.7$ & N/A\\
        \hline
        All time series & $\mathbf{91.9 \pm 0.7}$ & $84.3 \pm 0.8$ & $57.8 \pm 0.2$ &{\color{gray}  $50.6 \pm 0.6$} & N/A\\
        \hline
        Histogrammed distances & N/A & $\mathbf{76.8 \pm 0.9}$ & $75.1 \pm 0.7$ & $62.5 \pm 0.7$ & N/A\\
        \hline
        Histogrammed speeds & N/A & $\mathbf{72.0 \pm 0.8}$ & $70.9 \pm 1.0$ & $67.5 \pm 1.3$ & N/A\\
        \hline
    \end{tabular}
    \caption{Accuracy and its standard deviation (5-fold cross validation) of all trained models and input data. N/A indicates impossible combinations of input and classifier. Models within two standard deviations of chance $50\%$ are greyed out.
    The top-performing model for each feature set is highlighted in bold, the top performing models have a shaded table cell.}
    \label{tab:5_fold_accuracies}
\end{table}

The results presented in Fig.~\ref{fig:model_results}A show that the presence or lack of an attention layer has very little impact on MIDAS’s performance, and so does having the target object’s position. Moreover, it can be seen that MIDAS generalizes well to unseen users, even if only the raw gaze data is available (Fig.~\ref{fig:model_results}A). To better understand the performance of MIDAS over different lengths of input data streams, in Fig.~\ref{fig:model_results}B we present the evolution of MIDAS’s accuracy with respect to the time series duration it is being given to infer an intention. After only $600$ milliseconds, MIDAS can determine the user’s intent with $65\%$ accuracy and this only goes up with time, reaching a peak performance of around $92\%$ accuracy with $3.3$ seconds of input time series. This result shows that there is a latency versus accuracy trade-off when using MIDAS, which could potentially be set by users to suit their convenience. We see a similar trend for the best performing standard machine learning model, i.e. gradient boosting (Fig.~\ref{fig:model_results}B), showing an initial tie with MIDAS in shorter sequence lengths, with MIDAS taking over as sequence lengths increase. It must be noted that MIDAS has the advantage that a single model is trained to cope with time series of all lengths, whereas gradient boosting models need to retrain for different sequence lengths.

These results lead to two essential conclusions. First, we show the possibility of learning patterns in natural gaze behavior data which allows us to infer a users’ intent. Second, the longer the gaze sequence, the more accurate the decoding of the intention behind it. These two observations, confirmed by MIDAS and standard machine learning models as well, imply that there are characteristic patterns to specific intentions in our dataset of natural gaze behaviour, and that these cues are spread across time.

\section*{Discussion}
We set out to investigate whether we could utilise deep recurrent neural networks to overcome  the Midas touch problem of intention decoding in hand-based manipulations in a real-world, desk-based scenario.
Our deep classifier, MIDAS, learns to distinguish between different gaze dynamics in  real-world behaviour, allowing us to overcome the Midas Touch Problem\cite{velichkovsky1997towards} with a high degree of accuracy to distinguish between mere visual inspection and hand manipulation intentions. MIDAS can infer user intent with relatively little data, with the accuracy rising quickly from the first few hundred millisecond (including processing delays of verbal instruction) at $65\%$ to plateauing with $92\%$ at $3.3$ seconds (see Fig.~\ref{fig:model_results}B). This performance is in contrast to standard machine learning methods including methods developed previously to help resolve the Midas Touch problem.

We investigated whether the model was simply learning large spatial or temporal differences in the data. The gaze distance covered during the task varies considerably more across individuals than across the two classes, although there is a slight but systematic increase in gaze distance covered in the inspection task versus the manipulation task, see Fig.~\ref{fig:dataset_oveview}. This is perhaps to be expected, given the more active information-gathering for object comparisons that needs to take place in our inspection tasks, compared with a more focused approach in the manipulation tasks. There is considerable overlap between the two classes with respect to their spatial distributions, with the natural central-biased 2D Gaussians of both fitting with gaze behaviour described in the literature\cite{Tatler2007-gr}. Gaze speed is consistent between the two classes, with both classes' speeds fitting a chi-squared distribution closely, see Fig.~\ref{fig:dataset_oveview}. We investigated the effect of these potential biases within gaze-target distance, and gaze speed by training conventional machine learning algorithms on these features. The results (see Fig.~\ref{fig:model_results}C) show that MIDAS outperforms all of these models, suggesting that MIDAS is capturing dynamics beyond gaze-target distance and gaze speed.

A body of literature show how gaze temporal dynamics implicitly hold information both about intentional state (e.g. \cite{Pelz2001-ev, Iqbal2004-vz}) and task structure (e.g. \cite{Lengyel2021-af}. The weak performance of the GMM model indicates that gaze locations alone do not carry sufficient information to build reliable inverse Yarbus models. It is important to investigate the evolution of temporal structure across these natural tasks, if we are to properly build predictive models of natural behaviour that are consistent with the dynamic and ever-changing nature of our top-down goals. There is an unresolved question in the field as to what extent eye movements are driven by bottom-up processes \cite{Tatler2008-rs, Einhauser2008-px}, in response to visual features in a scene, versus by top-down processes, by our intentions and task goals. Our results with MIDAS lend weight to the combination of the two, but highlight the importance of evolving top-down subgoals in determining gaze location and future scanpaths.

Our approach uses unconstrained  natural gaze behaviour as a action intention signal from free head and body movement based eyetrscking in a real-world  setting involving physical manipulation of ovjects. We can now distnguish the subtle differences within eye movement trajectories across two major modes of intention that triggers eye movements.
Taking our approach into practice means,e.g. as a user interface in VR or as an asstivie technology for motor impaired,  we can implement a Zero User interface  (Zero UI) system \cite{subramanian2021gaze}, that detects the user's intentions while they act naturally, and will therefore not require any interactions with a user interface nor users having to learn custom gaze ``gestures'' (such as blink timings, dwell times, or complex eye movement patterns) to resolve the Midas Touch problem. MIDAS operates and is trained on natural eye movements occurring during real-world interactions. Our approach is entirely data-driven, and based on a new machine learning model that only requires free movement eye-tracking data combined with ego-centric video to operate. This means that our approach is customisable to many real-world, natural tasks, simply by including further scenarios in the training phase. We have previously shown how navigation intentions can be successfully decoded from natural eye movements with a Zero UI approach and deployed it successfully in wheel chair driving in real-world settings\cite{subramanian2019eye}. However, while the system there opeated on natural eye movement behaviour, the temporal component was less important in the navigational setting of a wheelchair. Here, we demonstrate the ability to resolve the Midas Touch problem for actual touch and physical manipulation, where we show that the temporal component of eye movements are essential for intention decoding.

We have shown a data-driven route to resolve the Midas Touch problem and decode natural eye movements associated with action intentions versus visual inspection, previously in navigation tasks\cite{subramanian2019eye,subramanian2021gaze} and now here in manual ones. Our findings prompt the question whether a general-purpose, multi-task, gaze-based intention decoder robust to the Midas Touch problem can be created simply through collecting sufficiently rich and diverse task data. The tools for ubiquitous data collection using wearable eye-tracking and mobile computing are there. The scale of the required labelling of the action intention could be a challenge if done by hand,  but automatic labelling for action intention either through automatic action recognition \cite{thomik2013real} or latent embeddings of movements of the body and the eyes \cite{xiloyannis2017gaussian, harston2021prediction} may be timely solutions. It has now been 60 years since yarbus, the rcognittion that eye movement interacxes have benefits with limitations due to the Midas Touch problem\cite{jacob_what_1990}. Resolving the Midas Touch problem using our approach may help further boost the deployment of eyetracking technologies from controlled lab spaces into e.g. inside vehicles to support consumers\cite{faisal2021predicting}, to non-invasive easy to setup-and-go assistive technology for motor impaired users\cite{shafti_robot_arm_2019}, and to create more intuitive human-computer and human-robot interaction settings. 

\section*{Methods}
\subsection*{Experimental setup}

\emph{Experimental design -} For our data collection experiments, we designed an experimental setup to collect natural gaze data for two specific intentions: Inspection and Manipulation. Figure~\ref{fig:data_collection} gives an overview of the experimental pipeline. Participants sit in front of a dining table with six different objects on it. The experiment was split into trials. A trial would start with the subject hearing a computer generated audio description of the task through an over-ear headset they were wearing. The subject then proceeds to fulfil the given task, after which the trial is complete and a new one begins. Trials were designed to capture the two different intentions when looking at an object, i.e. Inspection or Manipulation. 

The tasks in the inspection class relate to comparing objects on a given feature, e.g. \textit{``Weight, which is higher: A - Bottle, or B - Bowl?''}. The manipulation class of tasks involved different types of object manipulation, either requiring for the task to be physically performed, or imagined, e.g.
\textit{``Pick and put back: Banana.''} or 
\textit{``Imagine finger positions to eat from: Bowl.''}. A full list of given tasks are provided in supplementary materials. As observed in above examples, instructions are given in such a way so as to reveal the target object of interaction as the last word. The participants would therefore have to wait for the instructions to be finished before proceeding to complete the task. This enables us to control which part of the trial carries task-relevant gaze cues for our classification model to be trained on. Trials were taken by participants in sessions of one hour. In total, 16 subjects, all right-handed, took 2 to 3 sessions each, leading to a grand total of $13,679$ trials and about $45$ hours of recorded data.

\textit{Setup -} An experiment runner software was designed and coded for these experiments, which would handle the progression of experiments, including randomisation of tasks, and generation of voice commands, as well as recording all the streams of data and producing the final dataset for our classifier. The software was run on a laptop using an NVIDIA GTX 1080 with 64GB RAM. The participants wore a pair of over-ear headphones and eye-tracking glasses (SMI ETG 2W, SensoMotoric Instruments GmbH, Teltow, Germany). Optical proximity sensors were integrated into the dining table setup with an Arduino Uno which was then interfaced with by the experiment runner software via USB, allowing participants to indicate an answer in the comparison tasks by holding their hand over the respective sensor, to select A or B, and also keeping track of object places, i.e. which object space on the table is empty. Egocentric video and gaze position on that video were recorded from the eye-tracking glasses. Gaze position was recorded at an average frequency of $120~$Hz while video frame recordings are at $30~$Hz. As the different recording devices had different recording rates, the time code of each of their recordings were also saved to ensure proper reconstruction of the events' timeline when working with the dataset later on. The recorded images were then fed into Semantic Fovea \cite{auepanwiriyakul2018semantic_fovea} for object recognition and bounding box definition. The sensor values, as well as the given task and objects involved are also recorded. In summary, $3$ main classification-relevant assets were recorded: ego-centric video, ego-centric gaze position and object names and bounding boxes.

\subsection*{Data preprocessing, bias mitigation and model verification}

Gaze position data was used raw, except for the very few missing data points which were set to $0$. Object position was recorded at around $30Hz$ while gaze positions are at around $120Hz$. It was therefore necessary to interpolate the object positions to fit the recording rate of the gaze traces. There are three major reasons for gaps in object positions' time series: the difference in recording frequencies, a failure of the object detection algorithm and the object actually not being in the frame. An interpolation strategy was put in place to decide which gaps should be filled in and which ones should be left empty. The decision criterion was the length of the gap, consider that if data is missing due to a failure of the detection algorithm, the gap is likely to be smaller than the case if the object is simply not in the frame. From the CDF of object position gap sizes presented in Supplementary Fig.~\ref{fig:object_missing_data}, it was decided to interpolate all gaps of less than 100 data points ($\simeq 0.8$ seconds).

With the experimental design above, we anticipate two potential sources of bias: (1) Inspection task descriptions are longer than Manipulation task descriptions. (2) Two or three objects are named in Inspection tasks against a single object in Manipulation ones. We studied these potential biases and mitigated where necessary.

The first major difference between two trials with a different intention is their length. Indeed, instruction texts for inspection trials are longer on average (therefore longer to pronounce) than those with manipulation intention. As such, it could be possible to get a trial's corresponding intention solely from its length. To mitigate this bias, we only considered data which was recorded after the audio task description ended. However, even when considering only the post-instruction section of a trial, not all trials were the same length, still representing a bias. Fig.~\ref{fig:trial_length_plots}C presents the inverse trial lengths CDF for each intention class. The difference in trial lengths start to appear after 400 points of recording so the rest of the trials was discarded to present from any bias. We thus cropped each sample to keep only a central part from the end of the audio description to the point where most recordings stop. This left us with around $3.33$ seconds of recording for each trial to train our classifier on. See the effect of cropping the trials in Fig.~\ref{fig:trial_length_plots}.

The second bias which was studied comes from the fact that the number of objects named in the task descriptions was different with the associated intention. More precisely, two objects were named in inspection tasks against only one in manipulation tasks. This difference could impact the gaze trace by making the eyes move further from the target object during inspection trials. To check that the impact of this difference was light, we trained classifiers only on distance features. Fig.~\ref{fig:model_results}.b. shows that these models couldn't perform well at the task of intention decoding, therefore the bias was considered as not significant.

\subsection*{MIDAS and other classifiers}
All classification models mentioned in this paper were trained on the gaze and target object traces or a function of these (e.g. target-object distance, histograms, ... ). Models were trained on different variants of cross-validation. Cross-validation consists in splitting a dataset into a given number $N$ of folds, setting a specific fold as test set, training the model on all the $N-1$ others and do this for each fold. This allows to get $N$ samples of the model accuracy, giving an idea of the distribution of accuracies for a specific model. In this paper, two types of cross-validation are used. In $k$-fold cross-validation, the original dataset is split randomly into $k$ equally sized smaller sets: the folds. In leave one subject out, each fold is the data of a specific subject. 

All the standard models were trained with Python's \texttt{sklearn} and MIDAS was implemented using \texttt{Tensorflow 2}. All instances of MIDAS waas trained during 200 epochs. In details, MIDAS is composed of $2$ bidirectional LSTM layers with $256$ neurons each, one attention layer leading to $3$ dense layers with respectively $32$, $16$ and $2$ neurons, with ReLU activation for the first two and sigmoid activation on the output layer. MIDAS was trained to minimise a binary cross-entropy loss using the Adam optimiser with a learning rate of $2.5 \times 10^{-4}$.

\bibliography{references}

\begin{thebibliography}{10}
\urlstyle{rm}
\expandafter\ifx\csname url\endcsname\relax
  \def\url#1{\texttt{#1}}\fi
\expandafter\ifx\csname urlprefix\endcsname\relax\def\urlprefix{URL }\fi
\expandafter\ifx\csname doiprefix\endcsname\relax\def\doiprefix{DOI: }\fi
\providecommand{\bibinfo}[2]{#2}
\providecommand{\eprint}[2][]{\url{#2}}

\bibitem{Laurel1990}
\bibinfo{author}{(ed.) Laurel, B.}
\newblock \bibinfo{journal}{\bibinfo{title}{The art of human-computer interface
  design}}.
\newblock {\emph{\JournalTitle{New York: Addison-Wesley.}}}
  (\bibinfo{year}{1990}).

\bibitem{abbott_ultra-low-cost_2012}
\bibinfo{author}{Abbott, W.} \& \bibinfo{author}{Faisal, A.}
\newblock \bibinfo{journal}{\bibinfo{title}{Ultra-low-cost 3d gaze estimation:
  an intuitive high information throughput compliment to direct brain–machine
  interfaces}}.
\newblock {\emph{\JournalTitle{Journal of neural engineering}}}
  \textbf{\bibinfo{volume}{9}}, \bibinfo{pages}{046016} (\bibinfo{year}{2012}).

\bibitem{adjouadi2004remote}
\bibinfo{author}{Adjouadi, M.}, \bibinfo{author}{Sesin, A.},
  \bibinfo{author}{Ayala, M.} \& \bibinfo{author}{Cabrerizo, M.}
\newblock \bibinfo{title}{Remote eye gaze tracking system as a computer
  interface for persons with severe motor disability}.
\newblock In \emph{\bibinfo{booktitle}{International conference on computers
  for handicapped persons}}, \bibinfo{pages}{761--769}
  (\bibinfo{organization}{Springer}, \bibinfo{year}{2004}).

\bibitem{chin2008integrated}
\bibinfo{author}{Chin, C.~A.}, \bibinfo{author}{Barreto, A.},
  \bibinfo{author}{Cremades, J.~G.} \& \bibinfo{author}{Adjouadi, M.}
\newblock \bibinfo{journal}{\bibinfo{title}{Integrated electromyogram and
  eye-gaze tracking cursor control system for computer users with motor
  disabilities}}.
\newblock {\emph{\JournalTitle{Journal of Rehabilitation Research \&
  Development}}}  (\bibinfo{year}{2008}).

\bibitem{velichkovsky1997towards}
\bibinfo{author}{Velichkovsky, B.}, \bibinfo{author}{Sprenger, A.} \&
  \bibinfo{author}{Unema, P.}
\newblock \bibinfo{title}{Towards gaze-mediated interaction: Collecting
  solutions of the “midas touch problem”}.
\newblock In \emph{\bibinfo{booktitle}{Human-Computer Interaction
  INTERACT’97}}, \bibinfo{pages}{509--516} (\bibinfo{organization}{Springer},
  \bibinfo{year}{1997}).

\bibitem{Niehorster2020}
\bibinfo{author}{Niehorster, D.~C.} \emph{et~al.}
\newblock \bibinfo{journal}{\bibinfo{title}{The impact of slippage on the data
  quality of head-worn eye trackers}}.
\newblock {\emph{\JournalTitle{Behavior Research Methods 2019 52:3}}}
  \textbf{\bibinfo{volume}{52}}, \bibinfo{pages}{1140--1160},
  \url{10.3758/S13428-019-01307-0} (\bibinfo{year}{2020}).

\bibitem{Velichkovsky2014-da}
\bibinfo{author}{Velichkovsky, B.~B.}, \bibinfo{author}{Rumyantsev, M.~A.} \&
  \bibinfo{author}{Morozov, M.~A.}
\newblock \bibinfo{journal}{\bibinfo{title}{New solution to the midas touch
  problem: Identification of visual commands via extraction of focal
  fixations}}.
\newblock {\emph{\JournalTitle{Procedia computer science}}}
  \textbf{\bibinfo{volume}{39}}, \bibinfo{pages}{75--82}
  (\bibinfo{year}{2014}).

\bibitem{Buswell1935-hf}
\bibinfo{author}{Buswell, G.~T.}
\newblock \emph{\bibinfo{title}{How people look at pictures: a study of the
  psychology and perception in art}} (\bibinfo{publisher}{University of Chicago
  Press}, \bibinfo{year}{1935}).

\bibitem{yarbus1967eye}
\bibinfo{author}{Yarbus, A.~L.}
\newblock \bibinfo{title}{Eye movements during perception of complex objects}.
\newblock In \emph{\bibinfo{booktitle}{Eye movements and vision}},
  \bibinfo{pages}{171--211} (\bibinfo{publisher}{Springer},
  \bibinfo{year}{1967}).

\bibitem{grasso_eye-head_1998}
\bibinfo{author}{Grasso, R.}, \bibinfo{author}{Prévost, P.},
  \bibinfo{author}{Ivanenko, Y.~P.} \& \bibinfo{author}{Berthoz, A.}
\newblock \bibinfo{journal}{\bibinfo{title}{Eye-head coordination for the
  steering of locomotion in humans: an anticipatory synergy}}.
\newblock {\emph{\JournalTitle{Neuroscience Letters}}}
  \textbf{\bibinfo{volume}{253}}, \bibinfo{pages}{115--118}
  (\bibinfo{year}{1998}).

\bibitem{hollands_look_2002}
\bibinfo{author}{Hollands, M.~A.}, \bibinfo{author}{Patla, A.~E.} \&
  \bibinfo{author}{Vickers, J.~N.}
\newblock \bibinfo{journal}{\bibinfo{title}{“look where you’re going!”:
  gaze behaviour associated with maintaining and changing the direction of
  locomotion}}.
\newblock {\emph{\JournalTitle{Experimental brain research}}}
  \textbf{\bibinfo{volume}{143}}, \bibinfo{pages}{221--230}
  (\bibinfo{year}{2002}).

\bibitem{Becker1969-yj}
\bibinfo{author}{Becker, W.} \& \bibinfo{author}{Fuchs, A.~F.}
\newblock \bibinfo{journal}{\bibinfo{title}{Further properties of the human
  saccadic system: eye movements and correction saccades with and without
  visual fixation points}}.
\newblock {\emph{\JournalTitle{Vision research}}} \textbf{\bibinfo{volume}{9}},
  \bibinfo{pages}{1247--1258} (\bibinfo{year}{1969}).

\bibitem{Fuchs1976-vf}
\bibinfo{author}{Fuchs, A.~F.}
\newblock \bibinfo{journal}{\bibinfo{title}{The neurophysiology of saccades}}.
\newblock {\emph{\JournalTitle{Eye movements and psychological processes}}}
  \bibinfo{pages}{39--53} (\bibinfo{year}{1976}).

\bibitem{Rayner1992-tv}
\bibinfo{author}{Rayner, K.}
\newblock \bibinfo{title}{Eye movements and visual cognition: Introduction}.
\newblock In \bibinfo{editor}{Rayner, K.} (ed.) \emph{\bibinfo{booktitle}{Eye
  Movements and Visual Cognition: Scene Perception and Reading}},
  \bibinfo{pages}{1--7} (\bibinfo{publisher}{Springer New York},
  \bibinfo{address}{New York, NY}, \bibinfo{year}{1992}).

\bibitem{jacob_what_1990}
\bibinfo{author}{Jacob, R.~J.}
\newblock \bibinfo{title}{What you look at is what you get: eye movement-based
  interaction techniques}.
\newblock In \emph{\bibinfo{booktitle}{Proceedings of the {SIGCHI} conference
  on Human factors in computing systems}}, \bibinfo{pages}{11--18}
  (\bibinfo{year}{1990}).

\bibitem{salvucci2000identifying}
\bibinfo{author}{Salvucci, D.~D.} \& \bibinfo{author}{Goldberg, J.~H.}
\newblock \bibinfo{title}{Identifying fixations and saccades in eye-tracking
  protocols}.
\newblock In \emph{\bibinfo{booktitle}{Proceedings of the 2000 symposium on Eye
  tracking research \& applications}}, \bibinfo{pages}{71--78}
  (\bibinfo{year}{2000}).

\bibitem{duchowski_breadth-first_2002}
\bibinfo{author}{Duchowski, A.~T.}
\newblock \bibinfo{journal}{\bibinfo{title}{A breadth-first survey of
  eye-tracking applications}}.
\newblock {\emph{\JournalTitle{Behavior Research Methods, Instruments, \&
  Computers}}} \textbf{\bibinfo{volume}{34}}, \bibinfo{pages}{455--470}
  (\bibinfo{year}{2002}).
\newblock \bibinfo{note}{Publisher: Springer}.

\bibitem{Lorenz2016-es}
\bibinfo{author}{Lorenz, R.} \emph{et~al.}
\newblock \bibinfo{journal}{\bibinfo{title}{The automatic neuroscientist: A
  framework for optimizing experimental design with closed-loop real-time
  {fMRI}}}.
\newblock {\emph{\JournalTitle{NeuroImage}}} \textbf{\bibinfo{volume}{129}},
  \bibinfo{pages}{320--334} (\bibinfo{year}{2016}).

\bibitem{Rothkopf2004-wt}
\bibinfo{author}{Rothkopf, C.~A.} \& \bibinfo{author}{Pelz, J.~B.}
\newblock \bibinfo{journal}{\bibinfo{title}{Head movement estimation for
  wearable eye tracker}}.
\newblock {\emph{\JournalTitle{Proceedings of the 2004 symposium on eye}}}
  (\bibinfo{year}{2004}).

\bibitem{land_making_tea_1999}
\bibinfo{author}{Land, M.}, \bibinfo{author}{Mennie, N.} \&
  \bibinfo{author}{Rusted, J.}
\newblock \bibinfo{journal}{\bibinfo{title}{The roles of vision and eye
  movements in the control of activities of daily living}}.
\newblock {\emph{\JournalTitle{Perception}}} \textbf{\bibinfo{volume}{28}},
  \bibinfo{pages}{1311--1328} (\bibinfo{year}{1999}).
\newblock \bibinfo{note}{Publisher: SAGE Publications Sage UK: London,
  England}.

\bibitem{Hayhoe2003-be}
\bibinfo{author}{Hayhoe, M.~M.}, \bibinfo{author}{Shrivastava, A.},
  \bibinfo{author}{Mruczek, R.} \& \bibinfo{author}{Pelz, J.~B.}
\newblock \bibinfo{journal}{\bibinfo{title}{Visual memory and motor planning in
  a natural task}}.
\newblock {\emph{\JournalTitle{Journal of vision}}}
  \textbf{\bibinfo{volume}{3}}, \bibinfo{pages}{6} (\bibinfo{year}{2003}).

\bibitem{keshava2021just}
\bibinfo{author}{Keshava, A.} \emph{et~al.}
\newblock \bibinfo{journal}{\bibinfo{title}{Just-in-time: gaze guidance
  behavior while action planning and execution in vr}}.
\newblock {\emph{\JournalTitle{bioRxiv}}}  (\bibinfo{year}{2021}).

\bibitem{Johansson2001-mj}
\bibinfo{author}{Johansson, R.~S.}, \bibinfo{author}{Westling, G.},
  \bibinfo{author}{B{\"a}ckstr{\"o}m, A.} \& \bibinfo{author}{Flanagan, J.~R.}
\newblock \bibinfo{journal}{\bibinfo{title}{{Eye--Hand} coordination in object
  manipulation}}.
\newblock {\emph{\JournalTitle{The Journal of neuroscience: the official
  journal of the Society for Neuroscience}}} \textbf{\bibinfo{volume}{21}},
  \bibinfo{pages}{6917--6932} (\bibinfo{year}{2001}).

\bibitem{Haji-Abolhassani2014-gf}
\bibinfo{author}{Haji-Abolhassani, A.} \& \bibinfo{author}{Clark, J.~J.}
\newblock \bibinfo{journal}{\bibinfo{title}{An inverse yarbus process:
  predicting observers' task from eye movement patterns}}.
\newblock {\emph{\JournalTitle{Vision research}}}
  \textbf{\bibinfo{volume}{103}}, \bibinfo{pages}{127--142}
  (\bibinfo{year}{2014}).

\bibitem{Greene2012-ka}
\bibinfo{author}{Greene, M.~R.}, \bibinfo{author}{Liu, T.} \&
  \bibinfo{author}{Wolfe, J.~M.}
\newblock \bibinfo{journal}{\bibinfo{title}{Reconsidering yarbus: A failure to
  predict observers' task from eye movement patterns}}.
\newblock {\emph{\JournalTitle{Vision research}}}
  \textbf{\bibinfo{volume}{62}}, \bibinfo{pages}{1--8} (\bibinfo{year}{2012}).

\bibitem{Borji2014-py}
\bibinfo{author}{Borji, A.} \& \bibinfo{author}{Itti, L.}
\newblock \bibinfo{journal}{\bibinfo{title}{Defending yarbus: Eye movements
  reveal observers' task}}.
\newblock {\emph{\JournalTitle{Journal of vision}}}
  \textbf{\bibinfo{volume}{14}}, \bibinfo{pages}{29--29}
  (\bibinfo{year}{2014}).

\bibitem{Iqbal2004-vz}
\bibinfo{author}{Iqbal, S.~T.} \& \bibinfo{author}{Bailey, B.~P.}
\newblock \bibinfo{title}{Using eye gaze patterns to identify user tasks}.
\newblock
  \bibinfo{howpublished}{\url{https://www.interruptions.net/literature/Iqbal-GHC04.pdf}}
  (\bibinfo{year}{2004}).
\newblock \bibinfo{note}{Accessed: 2021-6-17}.

\bibitem{Henderson2013-oz}
\bibinfo{author}{Henderson, J.~M.}, \bibinfo{author}{Shinkareva, S.~V.},
  \bibinfo{author}{Wang, J.}, \bibinfo{author}{Luke, S.~G.} \&
  \bibinfo{author}{Olejarczyk, J.}
\newblock \bibinfo{journal}{\bibinfo{title}{Predicting cognitive state from eye
  movements}}.
\newblock {\emph{\JournalTitle{PloS one}}} \textbf{\bibinfo{volume}{8}},
  \bibinfo{pages}{e64937} (\bibinfo{year}{2013}).

\bibitem{Castelhano2009-mq}
\bibinfo{author}{Castelhano, M.~S.}, \bibinfo{author}{Mack, M.~L.} \&
  \bibinfo{author}{Henderson, J.~M.}
\newblock \bibinfo{journal}{\bibinfo{title}{Viewing task influences eye
  movement control during active scene perception}}.
\newblock {\emph{\JournalTitle{Journal of vision}}}
  \textbf{\bibinfo{volume}{9}}, \bibinfo{pages}{6.1--15}
  (\bibinfo{year}{2009}).

\bibitem{Bulling2013-qv}
\bibinfo{author}{Bulling, A.}, \bibinfo{author}{Weichel, C.} \&
  \bibinfo{author}{Gellersen, H.}
\newblock \bibinfo{title}{{EyeContext}: recognition of high-level contextual
  cues from human visual behaviour}.
\newblock In \emph{\bibinfo{booktitle}{Proceedings of the {SIGCHI} Conference
  on Human Factors in Computing Systems}}, CHI '13, \bibinfo{pages}{305--308}
  (\bibinfo{publisher}{Association for Computing Machinery},
  \bibinfo{address}{New York, NY, USA}, \bibinfo{year}{2013}).

\bibitem{Boisvert2016-my}
\bibinfo{author}{Boisvert, J. F.~G.} \& \bibinfo{author}{Bruce, N. D.~B.}
\newblock \bibinfo{journal}{\bibinfo{title}{Predicting task from eye movements:
  On the importance of spatial distribution, dynamics, and image features}}.
\newblock {\emph{\JournalTitle{Neurocomputing}}}
  \textbf{\bibinfo{volume}{207}}, \bibinfo{pages}{653--668}
  (\bibinfo{year}{2016}).

\bibitem{Hansen2003-su}
\bibinfo{author}{Hansen, J.~P.}, \bibinfo{author}{Johansen, A.~S.},
  \bibinfo{author}{Hansen, D.~W.}, \bibinfo{author}{Ito, K.} \&
  \bibinfo{author}{Mashino, S.}
\newblock \bibinfo{title}{Command without a click: Dwell time typing by mouse
  and gaze selections}.
\newblock In \emph{\bibinfo{booktitle}{{INTERACT}}}, vol.~\bibinfo{volume}{3},
  \bibinfo{pages}{121--128} (\bibinfo{publisher}{Citeseer},
  \bibinfo{year}{2003}).

\bibitem{Parisay2020-xw}
\bibinfo{author}{Parisay, M.}, \bibinfo{author}{Poullis, C.} \&
  \bibinfo{author}{Kersten, M.}
\newblock \bibinfo{journal}{\bibinfo{title}{{EyeTAP}: A novel technique using
  voice inputs to address the midas touch problem for gaze-based
  interactions}}.
\newblock {\emph{\JournalTitle{arxiv}}}  (\bibinfo{year}{2020}).
\newblock \eprint{2002.08455}.

\bibitem{dziemian2016gaze}
\bibinfo{author}{Dziemian, S.}, \bibinfo{author}{Abbott, W.~W.} \&
  \bibinfo{author}{Faisal, A.~A.}
\newblock \bibinfo{title}{Gaze-based teleprosthetic enables intuitive
  continuous control of complex robot arm use: Writing \& drawing}.
\newblock In \emph{\bibinfo{booktitle}{2016 6th IEEE International Conference
  on Biomedical Robotics and Biomechatronics (BioRob)}},
  \bibinfo{pages}{1277--1282} (\bibinfo{organization}{IEEE},
  \bibinfo{year}{2016}).

\bibitem{drone_GCI_2014}
\bibinfo{author}{Hansen, J.~P.}, \bibinfo{author}{Alapetite, A.},
  \bibinfo{author}{MacKenzie, I.~S.} \& \bibinfo{author}{Møllenbach, E.}
\newblock \bibinfo{title}{The use of gaze to control drones}.
\newblock In \emph{\bibinfo{booktitle}{Proceedings of the symposium on eye
  tracking research and applications}}, \bibinfo{pages}{27--34}
  (\bibinfo{year}{2014}).

\bibitem{tostado20163d}
\bibinfo{author}{Tostado, P.~M.}, \bibinfo{author}{Abbott, W.~W.} \&
  \bibinfo{author}{Faisal, A.~A.}
\newblock \bibinfo{title}{3d gaze cursor: Continuous calibration and end-point
  grasp control of robotic actuators}.
\newblock In \emph{\bibinfo{booktitle}{2016 IEEE International Conference on
  Robotics and Automation (ICRA)}}, \bibinfo{pages}{3295--3300}
  (\bibinfo{organization}{IEEE}, \bibinfo{year}{2016}).

\bibitem{Coutrot2018-jn}
\bibinfo{author}{Coutrot, A.}, \bibinfo{author}{Hsiao, J.~H.} \&
  \bibinfo{author}{Chan, A.~B.}
\newblock \bibinfo{journal}{\bibinfo{title}{Scanpath modeling and
  classification with hidden markov models}}.
\newblock {\emph{\JournalTitle{Behavior Research Methods}}}
  \textbf{\bibinfo{volume}{50}}, \bibinfo{pages}{362--379}
  (\bibinfo{year}{2018}).

\bibitem{auepanwiriyakul2018semantic_fovea}
\bibinfo{author}{Auepanwiriyakul, C.}, \bibinfo{author}{Harston, A.},
  \bibinfo{author}{Orlov, P.}, \bibinfo{author}{Shafti, A.} \&
  \bibinfo{author}{Faisal, A.~A.}
\newblock \bibinfo{title}{Semantic fovea: real-time annotation of ego-centric
  videos with gaze context}.
\newblock In \emph{\bibinfo{booktitle}{Proceedings of the 2018 ACM Symposium on
  Eye Tracking Research \& Applications}}, \bibinfo{pages}{1--3}
  (\bibinfo{year}{2018}).

\bibitem{Tatler2007-gr}
\bibinfo{author}{Tatler, B.~W.}
\newblock \bibinfo{journal}{\bibinfo{title}{The central fixation bias in scene
  viewing: Selecting an optimal viewing position independently of motor biases
  and image feature distributions}}.
\newblock {\emph{\JournalTitle{Journal of vision}}}
  \textbf{\bibinfo{volume}{7}}, \bibinfo{pages}{4} (\bibinfo{year}{2007}).

\bibitem{reynolds_GMM_2009}
\bibinfo{author}{Reynolds, D.~A.}
\newblock \bibinfo{journal}{\bibinfo{title}{Gaussian mixture models.}}
\newblock {\emph{\JournalTitle{Encyclopedia of biometrics}}}
  \textbf{\bibinfo{volume}{741}} (\bibinfo{year}{2009}).
\newblock \bibinfo{note}{Publisher: Springer City}.

\bibitem{neath_BIC_2012}
\bibinfo{author}{Neath, A.~A.} \& \bibinfo{author}{Cavanaugh, J.~E.}
\newblock \bibinfo{journal}{\bibinfo{title}{The {Bayesian} information
  criterion: background, derivation, and applications}}.
\newblock {\emph{\JournalTitle{Wiley Interdisciplinary Reviews: Computational
  Statistics}}} \textbf{\bibinfo{volume}{4}}, \bibinfo{pages}{199--203}
  (\bibinfo{year}{2012}).
\newblock \bibinfo{note}{Publisher: Wiley Online Library}.

\bibitem{land_tea_1999}
\bibinfo{author}{Land, M.}, \bibinfo{author}{Mennie, N.} \&
  \bibinfo{author}{Rusted, J.}
\newblock \bibinfo{journal}{\bibinfo{title}{The roles of vision and eye
  movements in the control of activities of daily living}}.
\newblock {\emph{\JournalTitle{Perception}}} \textbf{\bibinfo{volume}{28}},
  \bibinfo{pages}{1311--1328} (\bibinfo{year}{1999}).
\newblock \bibinfo{note}{Publisher: {SAGE} Publications Sage {UK}: London,
  England}.

\bibitem{rosenthal_semeval-2017_2017}
\bibinfo{author}{Rosenthal, S.}, \bibinfo{author}{Farra, N.} \&
  \bibinfo{author}{Nakov, P.}
\newblock \bibinfo{title}{{SemEval}-2017 task 4: {Sentiment} analysis in
  {Twitter}}.
\newblock In \emph{\bibinfo{booktitle}{Proceedings of the 11th international
  workshop on semantic evaluation (SemEval-2017)}}, \bibinfo{pages}{502--518}
  (\bibinfo{year}{2017}).

\bibitem{baziotis_datastories_2017}
\bibinfo{author}{Baziotis, C.}, \bibinfo{author}{Pelekis, N.} \&
  \bibinfo{author}{Doulkeridis, C.}
\newblock \bibinfo{title}{Datastories at semeval-2017 task 4: {Deep} lstm with
  attention for message-level and topic-based sentiment analysis}.
\newblock In \emph{\bibinfo{booktitle}{Proceedings of the 11th international
  workshop on semantic evaluation (SemEval-2017)}}, \bibinfo{pages}{747--754}
  (\bibinfo{year}{2017}).

\bibitem{graves2005framewise_bidirLSTM}
\bibinfo{author}{Graves, A.} \& \bibinfo{author}{Schmidhuber, J.}
\newblock \bibinfo{journal}{\bibinfo{title}{Framewise phoneme classification
  with bidirectional lstm and other neural network architectures}}.
\newblock {\emph{\JournalTitle{Neural networks}}}
  \textbf{\bibinfo{volume}{18}}, \bibinfo{pages}{602--610}
  (\bibinfo{year}{2005}).

\bibitem{bahdanau_neuralAttention_2014}
\bibinfo{author}{Bahdanau, D.}, \bibinfo{author}{Cho, K.} \&
  \bibinfo{author}{Bengio, Y.}
\newblock \bibinfo{journal}{\bibinfo{title}{Neural machine translation by
  jointly learning to align and translate}}.
\newblock {\emph{\JournalTitle{arXiv preprint arXiv:1409.0473}}}
  (\bibinfo{year}{2014}).

\bibitem{lin2014microsoft}
\bibinfo{author}{Lin, T.-Y.} \emph{et~al.}
\newblock \bibinfo{title}{Microsoft coco: Common objects in context}.
\newblock In \emph{\bibinfo{booktitle}{European conference on computer
  vision}}, \bibinfo{pages}{740--755} (\bibinfo{organization}{Springer},
  \bibinfo{year}{2014}).

\bibitem{deng2009imagenet}
\bibinfo{author}{Deng, J.} \emph{et~al.}
\newblock \bibinfo{title}{Imagenet: A large-scale hierarchical image database}.
\newblock In \emph{\bibinfo{booktitle}{2009 IEEE Conference on Computer Vision
  and Pattern Recognition (CVPR)}}, \bibinfo{pages}{248--255}
  (\bibinfo{organization}{Ieee}, \bibinfo{year}{2009}).

\bibitem{Pelz2001-ev}
\bibinfo{author}{Pelz, J.}, \bibinfo{author}{Hayhoe, M.} \&
  \bibinfo{author}{Loeber, R.}
\newblock \bibinfo{journal}{\bibinfo{title}{The coordination of eye, head, and
  hand movements in a natural task}}.
\newblock {\emph{\JournalTitle{Experimental brain research. Experimentelle
  Hirnforschung. Experimentation cerebrale}}} \textbf{\bibinfo{volume}{139}},
  \bibinfo{pages}{266--277} (\bibinfo{year}{2001}).

\bibitem{Lengyel2021-af}
\bibinfo{author}{Lengyel, G.}, \bibinfo{author}{Carlberg, K.},
  \bibinfo{author}{Samad, M.} \& \bibinfo{author}{Jonker, T.}
\newblock \bibinfo{journal}{\bibinfo{title}{Predicting visual attention using
  the hidden structure in eye-gaze dynamics}}.
\newblock {\emph{\JournalTitle{CHI EMICS 2021}}}  (\bibinfo{year}{2021}).

\bibitem{Tatler2008-rs}
\bibinfo{author}{Tatler, B.~W.} \& \bibinfo{author}{Vincent, B.~T.}
\newblock \bibinfo{journal}{\bibinfo{title}{Systematic tendencies in scene
  viewing}}.
\newblock {\emph{\JournalTitle{Journal of eye movement research}}}
  \textbf{\bibinfo{volume}{2}} (\bibinfo{year}{2008}).

\bibitem{Einhauser2008-px}
\bibinfo{author}{Einh{\"a}user, W.}, \bibinfo{author}{Spain, M.} \&
  \bibinfo{author}{Perona, P.}
\newblock \bibinfo{journal}{\bibinfo{title}{Recognition and attention: Relation
  of eye-position and object recall to bottom-up models of saliency}}.
\newblock {\emph{\JournalTitle{Journal of Vision}}}  (\bibinfo{year}{2008}).

\bibitem{subramanian2021gaze}
\bibinfo{author}{Subramanian, M.}, \bibinfo{author}{Park, S.},
  \bibinfo{author}{Orlov, P.}, \bibinfo{author}{Shafti, A.} \&
  \bibinfo{author}{Faisal, A.~A.}
\newblock \bibinfo{title}{Gaze-contingent decoding of human navigation
  intention on an autonomous wheelchair platform}.
\newblock In \emph{\bibinfo{booktitle}{2021 10th International IEEE/EMBS
  Conference on Neural Engineering (NER)}}, \bibinfo{pages}{335--338}
  (\bibinfo{organization}{IEEE}, \bibinfo{year}{2021}).

\bibitem{subramanian2019eye}
\bibinfo{author}{Subramanian, M.}, \bibinfo{author}{Songur, N.},
  \bibinfo{author}{Adjei, D.}, \bibinfo{author}{Orlov, P.} \&
  \bibinfo{author}{Faisal, A.~A.}
\newblock \bibinfo{title}{A. eye drive: Gaze-based semi-autonomous wheelchair
  interface}.
\newblock In \emph{\bibinfo{booktitle}{IEEE Engineering in Medicine and Biology
  (EMBC)}}, vol.~\bibinfo{volume}{41}, \bibinfo{pages}{5967--5970}
  (\bibinfo{organization}{IEEE}, \bibinfo{year}{2019}).

\bibitem{thomik2013real}
\bibinfo{author}{Thomik, A.~A.}, \bibinfo{author}{Haber, D.} \&
  \bibinfo{author}{Faisal, A.~A.}
\newblock \bibinfo{title}{Real-time movement prediction for improved control of
  neuroprosthetic devices}.
\newblock In \emph{\bibinfo{booktitle}{IEEE Neural Engineering (NER)}},
  vol.~\bibinfo{volume}{6}, \bibinfo{pages}{625--628}
  (\bibinfo{organization}{IEEE}, \bibinfo{year}{2013}).

\bibitem{xiloyannis2017gaussian}
\bibinfo{author}{Xiloyannis, M.}, \bibinfo{author}{Gavriel, C.},
  \bibinfo{author}{Thomik, A.~A.} \& \bibinfo{author}{Faisal, A.~A.}
\newblock \bibinfo{journal}{\bibinfo{title}{Gaussian process autoregression for
  simultaneous proportional multi-modal prosthetic control with natural hand
  kinematics}}.
\newblock {\emph{\JournalTitle{IEEE Transactions on Neural Systems and
  Rehabilitation Engineering}}} \textbf{\bibinfo{volume}{25}},
  \bibinfo{pages}{1785--1801} (\bibinfo{year}{2017}).

\bibitem{harston2021prediction}
\bibinfo{author}{Harston, J.~A.}, \bibinfo{author}{Auepanwiriyakul, C.} \&
  \bibinfo{author}{Faisal, A.}
\newblock \bibinfo{journal}{\bibinfo{title}{Prediction of visual attention in
  embodied real-world tasks}}.
\newblock {\emph{\JournalTitle{Journal of Vision}}}
  \textbf{\bibinfo{volume}{21}}, \bibinfo{pages}{2741--2741}
  (\bibinfo{year}{2021}).

\bibitem{faisal2021predicting}
\bibinfo{author}{Faisal, A.}
\newblock \bibinfo{journal}{\bibinfo{title}{Predicting visual attention of
  human drivers boosts the training speed and performance of autonomous
  vehicles}}.
\newblock {\emph{\JournalTitle{Journal of Vision}}}
  \textbf{\bibinfo{volume}{21}}, \bibinfo{pages}{2819--2819}
  (\bibinfo{year}{2021}).

\bibitem{shafti_robot_arm_2019}
\bibinfo{author}{Shafti, A.}, \bibinfo{author}{Orlov, P.} \&
  \bibinfo{author}{Faisal, A.~A.}
\newblock \bibinfo{title}{Gaze-based, context-aware robotic system for assisted
  reaching and grasping}.
\newblock In \emph{\bibinfo{booktitle}{2019 International Conference on
  Robotics and Automation ({ICRA})}}, \bibinfo{pages}{863--869}
  (\bibinfo{publisher}{{IEEE}}, \bibinfo{year}{2019}).

\end{thebibliography}

\section*{Supplementary Material}

% Resetting figures counter and adding S prefix.
\setcounter{figure}{0}
\makeatletter 
\renewcommand{\thefigure}{S\@arabic\c@figure}
\makeatother

\begin{figure}[htbp]
\centering
\includegraphics[width=.5\linewidth]{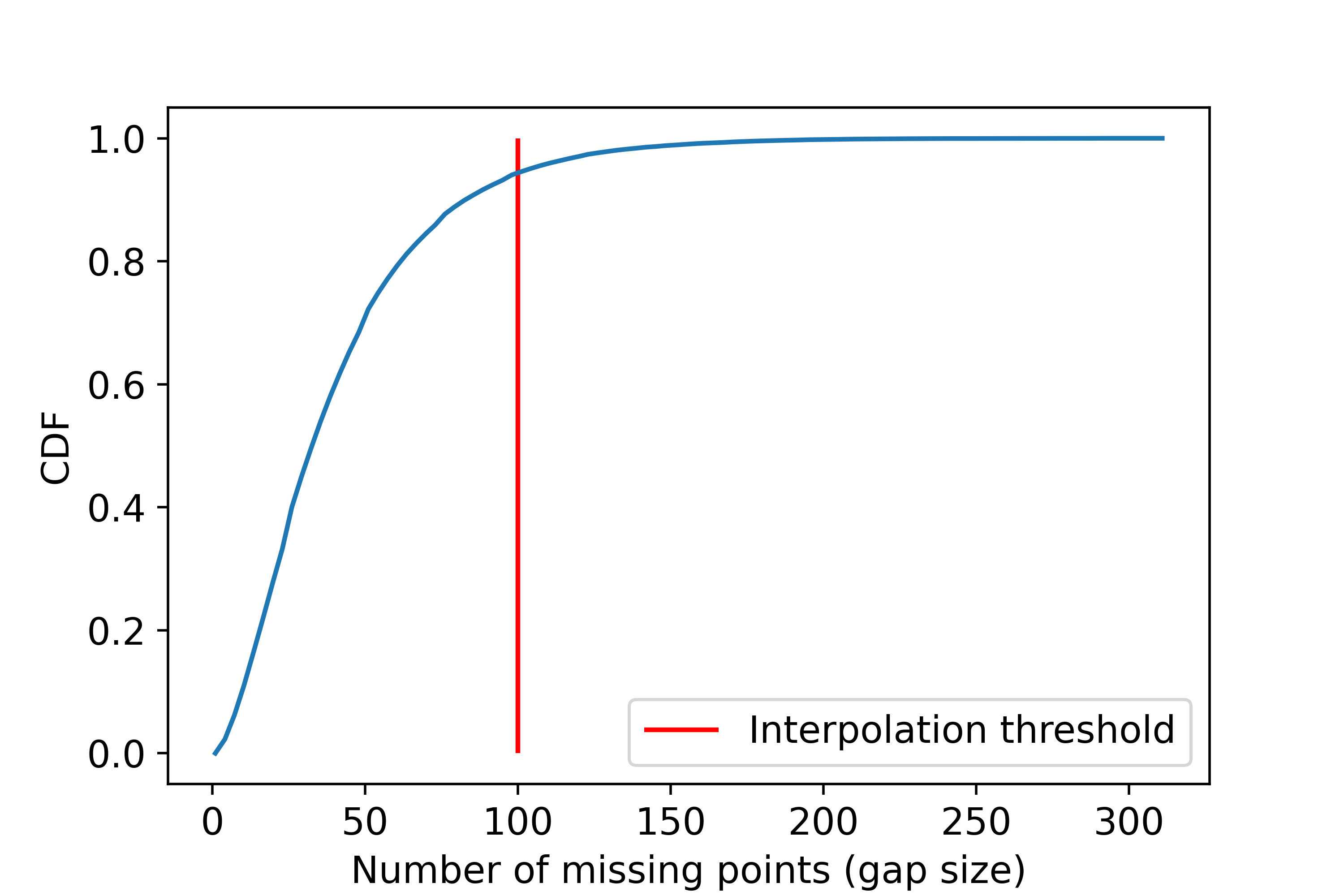}
\caption{Cumulative Density Function (CDF) of the distribution of the length of gaps in the object position time series.}
\label{fig:object_missing_data}
\end{figure}

\begin{figure}[htbp]
\centering
\includegraphics[width=\linewidth]{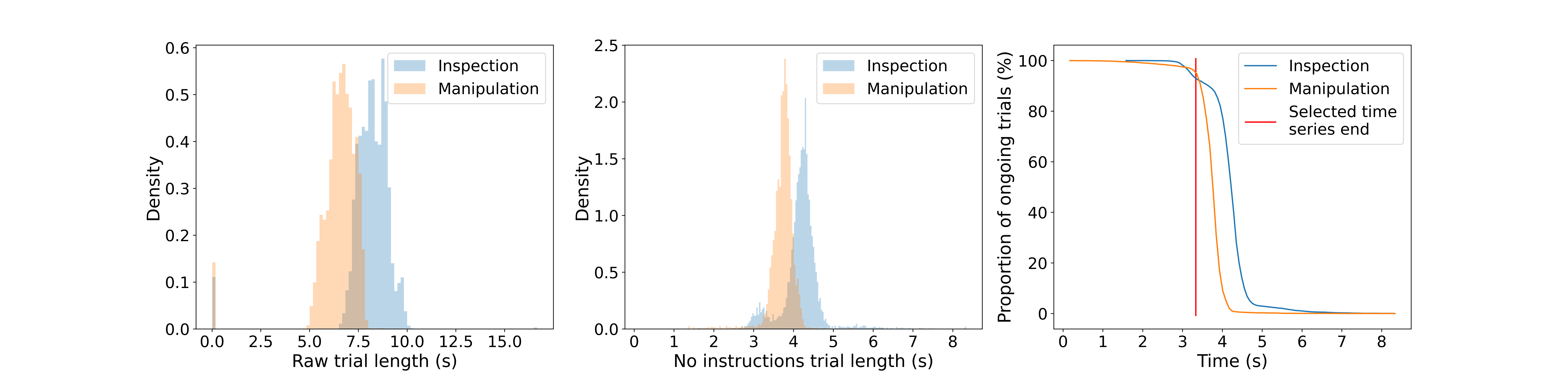}
\caption{Illustration of the effect of our preprocessing to mitigate the trial length bias. (a) shows the distribution of raw trial lengths (task description + task execution) for both intention classes.. (b) shows the distribution of task execution length for each trial (trial length minus instruction length) for both intention classes.. (c) shows the inverse CDF of the task execution lengths for both intention classes as well as the cutting point we used as end of input sequences.}
\label{fig:trial_length_plots}
\end{figure}

\end{document}